\documentclass[runningheads]{llncs}
\usepackage{graphicx}
\usepackage{times}
\usepackage{latexsym}
\usepackage{subfigure}
\usepackage{booktabs}
\usepackage{paralist}
\usepackage{bookmark,xspace}
\usepackage{multirow}
\usepackage{amsmath, amssymb}

\newcommand{\name}{RKDEA\xspace}
\usepackage{microtype}

\begin{document}
\title{Jointly Learning Knowledge Embedding and Neighborhood Consensus with Relational Knowledge Distillation for Entity Alignment}

\author{Xinhang Li \inst{1} \and Yong Zhang \inst{1}  \and Chunxiao Xing \inst{1} \\
  \texttt{xh-li20@mails.tsinghua.edu.cn} \\
  \texttt{zhangyong05@tsinghua.edu.cn} \\
  \texttt{xingcx@tsinghua.edu.cn}}
\authorrunning{Xinhang et al.}
\institute{Department of Computer Science, Tsinghua University, China}
\maketitle
\begin{abstract}
Entity alignment aims at integrating heterogeneous knowledge from different knowledge graphs.
Recent studies employ embedding-based methods by first learning the representation of Knowledge Graphs and then performing entity alignment via measuring the similarity between entity embeddings. 
However, they failed to make good use of the relation semantic information due to the trade-off problem caused by the different objectives of learning knowledge embedding and neighborhood consensus.
To address this problem, we propose \textbf{R}elational \textbf{K}nowledge \textbf{D}istillation for \textbf{E}ntity \textbf{A}lignment (\textbf{RKDEA}), a Graph Convolutional Network (GCN) based model equipped with knowledge distillation for entity alignment.
We adopt GCN-based models to learn the representation of entities by considering the graph structure and incorporating the relation semantic information into GCN via knowledge distillation.
Then, we introduce a novel adaptive mechanism to transfer relational knowledge so as to jointly learn entity embedding and neighborhood consensus.
Experimental results on several benchmarking datasets demonstrate the effectiveness of our proposed model.
\end{abstract}

\section{Introduction}
Knowledge Graphs (KGs) are able to provide unstructured knowledge in the simple and clear triple format \textit{\textless head, relation, tail\textgreater}.
They are essential in supporting many natural language processing applications.
Since KGs are constructed separately from heterogeneous resources and languages, they might use different expressions to indicate the same entity.
As a result, different KGs often contain complementary contents and cross-lingual links.
It is essential to integrate heterogeneous KGs into a unified one, thus increasing the accuracy and robustness of knowledge-driven applications.

\begin{figure}
    \centering
    \includegraphics[width=\linewidth]{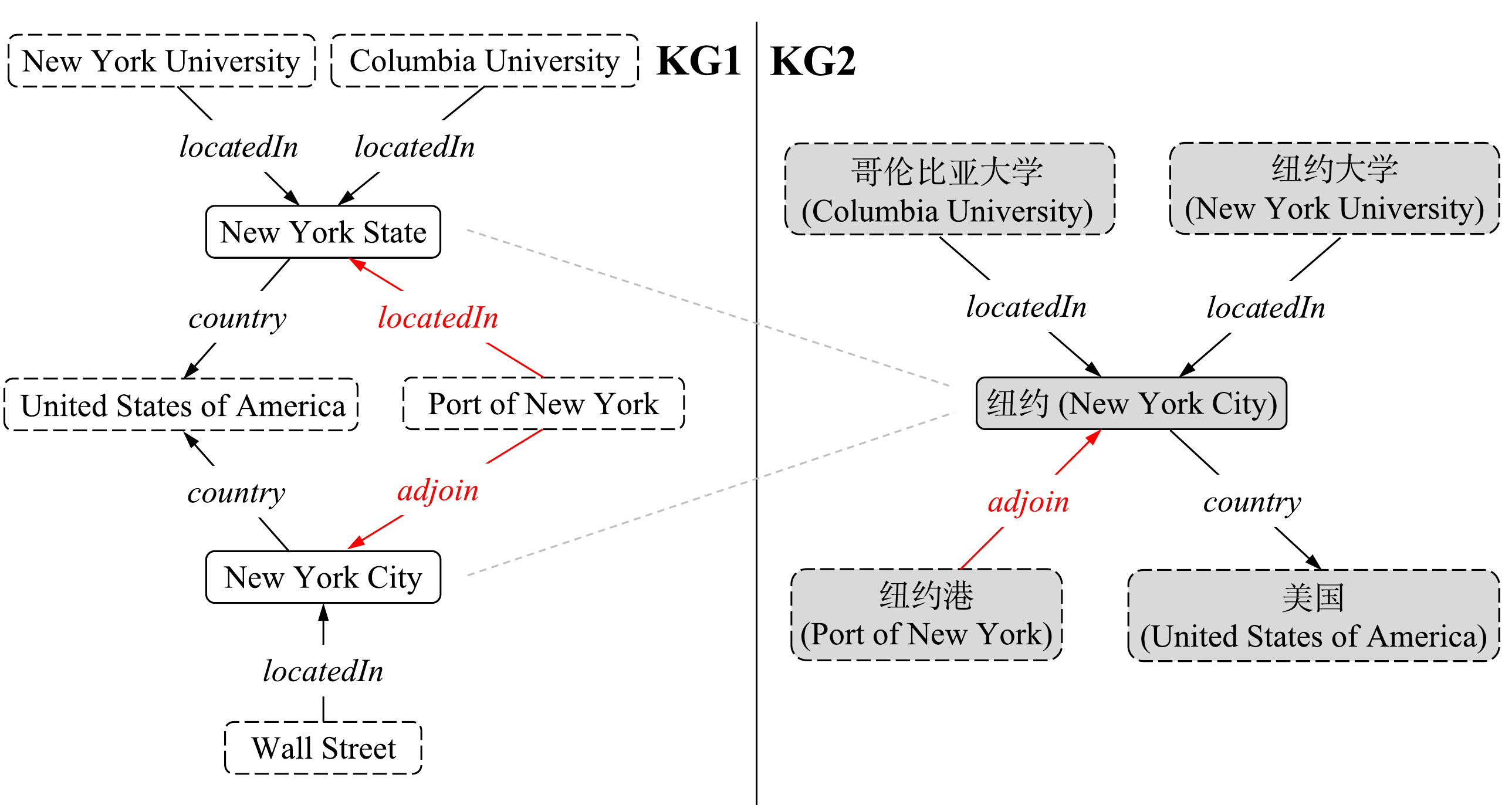}
    \caption{An error-prone example of two different entities with same neighbors.}\label{fig-motiex}
\end{figure}


To this end, many efforts have been paid to study the problem of Entity Alignment, which aims at linking entities with the same identities from different KGs.
Earlier approaches for entity alignment usually rely on manually created features~\cite{DBLP:conf/cidr/MahdisoltaniBS15}, which is labor-intensive and time-consuming.
Recent studies focused on embedding-based approaches as they are capable of representing and preserving the structures of KGs in low-dimensional embedding spaces.
Generally speaking, there are two categories of approaches: translation-based and GNN-based.
The translation-based models~\cite{DBLP:conf/semweb/SunHL17,DBLP:conf/ijcai/ZhuXLS17,DBLP:conf/ijcai/ChenTYZ17,DBLP:conf/ijcai/SunHZQ18} extend the idea of trans-family models, e.g. TransE, for knowledge graph embedding to learn the embedding of entities and relations in KGs.
This kind of method is good at learning knowledge embedding but not satisfactory in sparse graph entity alignment.
Recently GNN-based models~\cite{DBLP:conf/emnlp/WangLLZ18,DBLP:conf/esws/SchlichtkrullKB18} employ the Graph Convolutional Network (GCN)~\cite{DBLP:conf/iclr/KipfW17} to make better use of the pre-aligned seeds to learn the entity embedding by utilizing the neighbor information so as to resolve such limitations and have achieved promising results.
Some recent works~\cite{DBLP:conf/emnlp/LiCHSLC19} further jointly learned relational knowledge and neighborhood consensus~\cite{DBLP:conf/nips/RoccoCATPS18} to get more robust and accurate predictions.
However, since the learning objectives of relational knowledge and neighborhood consensus are different, it will lead to different optimization directions.
As a result, the model would fail to learn the useful information due to the overfitting problem.
For example, we can see in Figure~\ref{fig-motiex} that existing models tend to wrongly align \textit{New York State} in English and \textit{New York City} in Chinese due to the strong hint of neighborhood consensus given by the same neighbors.
However, in such a situation, the difference of relation semantic information between \textit{adjoin} and \textit{locatedIn} is more crucial in alignment to distinguish these two different entities, which cannot be treated properly without making a balance between the two learning objectives.


To address this problem, we propose \textbf{R}elational \textbf{K}nowledge \textbf{D}istillation for \textbf{E}ntity \textbf{A}lignment (\textbf{RKDEA}), a GCN based model with relational knowledge distillation framework for entity alignment.
Following previous studies~\cite{DBLP:conf/ijcai/WuLF0Y019,DBLP:conf/emnlp/WuLFWZ19,DBLP:conf/acl/WuLFWZ20,DBLP:conf/ijcai/NieHSWCWZ20}, we use a Highway Gated GCN model to learn the entity embedding.
To decide the portion of relational knowledge embedding and neighborhood consensus in the training objective, we take advantage of the \emph{knowledge distillation}~\cite{DBLP:journals/corr/HintonVD15} mechanism.
More specifically, we first separately train two models with objectives of learning relational knowledge and neighborhood consensus.
Next, we take the model with relational knowledge objective as teacher and the model with neighborhood consensus objective as student.
Then we employ a relational distillation method to transfer relation information from the teacher model to the student.
To effectively control the overall training objectives, we propose an adaptive temperature mechanism instead of treating it as a static hyper-parameter as previous studies~\cite{DBLP:journals/corr/HintonVD15,DBLP:journals/corr/RomeroBKCGB14,DBLP:conf/iclr/ZagoruykoK17,DBLP:conf/cvpr/ParkKLC19} did to adjust the weight of two kinds of information.
We conduct extensive evaluations on several publicly available datasets that are widely used in previous studies.
The experimental results and further analysis demonstrate that \name can better integrate knowledge embedding and neighborhood consensus and thus outperforms state-of-the-art methods by an obvious margin.

\section{Related Work}
\subsection{Knowledge Graph Entity Alignment}

Knowledge Graph has a wide scope of application scenarios like similarity search~\cite{DBLP:journals/tkde/ZhangWWX20,DBLP:conf/icde/WangLZ19,DBLP:conf/icde/WuZWLFX19,DBLP:conf/cikm/LuLW019}, information extraction~\cite{DBLP:journals/isci/WangLLZ20}, record de-duplication~\cite{DBLP:journals/jdiq/LiLSWHT21}, and can help analyze different kinds of of data such as health~\cite{DBLP:conf/dasfaa/ZhaoZWYZWX18}, spatial~\cite{DBLP:journals/www/ZhangCYWHXZ21,DBLP:journals/tkde/LiuADZWH22,DBLP:conf/icde/YangZZWHX19} and text~\cite{DBLP:conf/ijcai/TianZWX19}. 
To automatically capture deep resemblance of graph structure information between heterogeneous KGs, recent studies focused on embedding-based approach.
Based on the methodology, they could be categorized into two types: translation-based and GNN-based ones.

\noindent\textbf{Translation-based Approaches}\hspace{.5em}
For the representation learning of a single KG, there have been many studies, such as TransE~\cite{DBLP:conf/nips/BordesUGWY13},TransH~\cite{DBLP:conf/aaai/WangZFC14}, TransR~\cite{DBLP:conf/aaai/LinLSLZ15} and TransD~\cite{DBLP:conf/acl/JiHXL015}.
These methods utilize a scoring function to model relational knowledge and therefore obtain entity and relation embeddings.
Translation-based approaches are based on such studies. 
MTransE~\cite{DBLP:conf/ijcai/ChenTYZ17} applies TransE into entity alignment task with various transition techniques between different KGs.
JAPE~\cite{DBLP:conf/semweb/SunHL17} presents a way to combine structure and attribute information to jointly embed entities into a unified vector space.
This kind of method is capable of capturing complex relation semantic information with the help of triple-level modeling.
However, it is difficult for them to perceive the structural similarity of neighborhood information.

\noindent\textbf{GNN-based Approaches}\hspace{.5em}
Recently, the Graph Neural Network(GNN) has achieved tremendous success on the applications related to network embedding.
GNN-based entity alignment methods incorporate neighborhood information with GNNs to provide global structural information.
GCN-Align~\cite{DBLP:conf/emnlp/WangLLZ18} directly applies the Graph Convolutional Network (GCN) as embedding module on entity alignment.
MuGNN~\cite{DBLP:conf/acl/CaoLLLLC19} proposes self and cross-KG attention mechanisms to better capture the structure information in the KGs.
RDGCN~\cite{DBLP:conf/ijcai/WuLF0Y019} leverages relations to improve entity alignment with dual graphs.
GMNN~\cite{DBLP:conf/acl/XuWYFSWY19} and NMN~\cite{DBLP:conf/acl/WuLFWZ20} incorporate long-distance neighborhood information to strengthen the entity embeddings.
SSP~\cite{DBLP:conf/ijcai/NieHSWCWZ20} jointly models KG global structure and local semantic information via flexible relation representation.
KECG~\cite{DBLP:conf/emnlp/LiCHSLC19} trains model with knowledge embedding and neighborhood consensus objectives alternately.

Nevertheless, all these methods fail to address the problem of balancing the different learning objectives of relational knowledge and neighborhood consensus.
Comparing with them, our approach could jointly learn knowledge embedding and neighborhood consensus in a more structured, fine-grained way via knowledge distillation.

\subsection{Knowledge Distillation}

Knowledge Distillation(KD) is a branch of transfer learning, which indicates transferring knowledge from a complex model (teacher) to a concise model (student).
Typically, KD aims at transferring a mapping from inputs to outputs learned by teacher model to student model.
By leveraging KD, the student model could learn implicit knowledge by incorporating an extra objective of the teacher's outputs so as to gain better performance.
It was first introduced to neural network by~\cite{DBLP:journals/corr/HintonVD15}.
\cite{DBLP:journals/corr/RomeroBKCGB14} employs additional linear transformation in the middle of network to get a narrower student.
\cite{DBLP:conf/iclr/ZagoruykoK17,DBLP:journals/corr/HuangW17a,DBLP:conf/nips/TarvainenV17} transfer the knowledge in attention map to get robuster and more comprehensive representations.
Recently, some works~\cite{DBLP:conf/cvpr/YimJBK17,DBLP:journals/corr/abs-1805-02641,DBLP:conf/icml/FurlanelloLTIA18} have demonstrated that distilling models of identical architecture, i.e., self-distillation, can further improve the performance of neural network.

The challenges for graph representation learning lies in the heterogeneous nature of different graphs.
Some recent approaches of applying KD have brought convincing results in solving heterogeneity problem.
\cite{DBLP:conf/aaai/LeeKLY17} proposes a novel graph data transfer learning framework with generalized Spectral CNN.
\cite{DBLP:conf/aaai/ChenWZ18} transfers similarity of different structure for metric learning.
\cite{DBLP:conf/cvpr/ParkKLC19} indicates the effectiveness of distance-wise and angle-wise distillation loss in knowledge transfer between different structures.
The objective of our work is similar to above studies but there are still many issues to be addressed to propose a reasonable distillation mechanism for the task of entity alignment between KGs.

\section{Preliminary}

Formally, a knowledge graph is defined as $G = \langle E, R, A, T_R, T_A \rangle$, where $E, R, A$ indicate sets of entities, relations and attributes, respectively.
$T_R \subset E \times R \times E $ and $T_A \subset E \times A \times V$ denote sets of relation triples and attribute triples, respectively.
In this paper, we focus on relation information irrespective of attributes.
So the KG could be simplified to $G = \langle E, R, T \rangle$, where $T$ is the set of relation triples. 

Given two heterogeneous KGs, $G_1 = \langle E_1, R_1, T_1 \rangle$ and $G_2 =  \langle E_2, R_2, T_2 \rangle$, Entity Alignment aims at finding entity pairs $\langle e_1, e_2\rangle$, $e_1 \in E_1$ and $e_2 \in E_2$ that represent same meaning semantically.
In practice, there are always some pre-aligned entity and relation pairs provided as \emph{seed alignment}.
The seed alignments $\mathcal{S}_E = \{(e_1, e_2) \in E_1 \times E_2 | e_1 \doteq e_2\}$ and $\mathcal{S}_R = \{(r_1, r_2) \in R_1 \times R_2 | r_1 \doteq r_2\}$ where $\doteq$ means equivalent in semantics, denote semantically equivalent pairs.

\begin{figure*}
  \centering 
  \includegraphics[width=\textwidth]{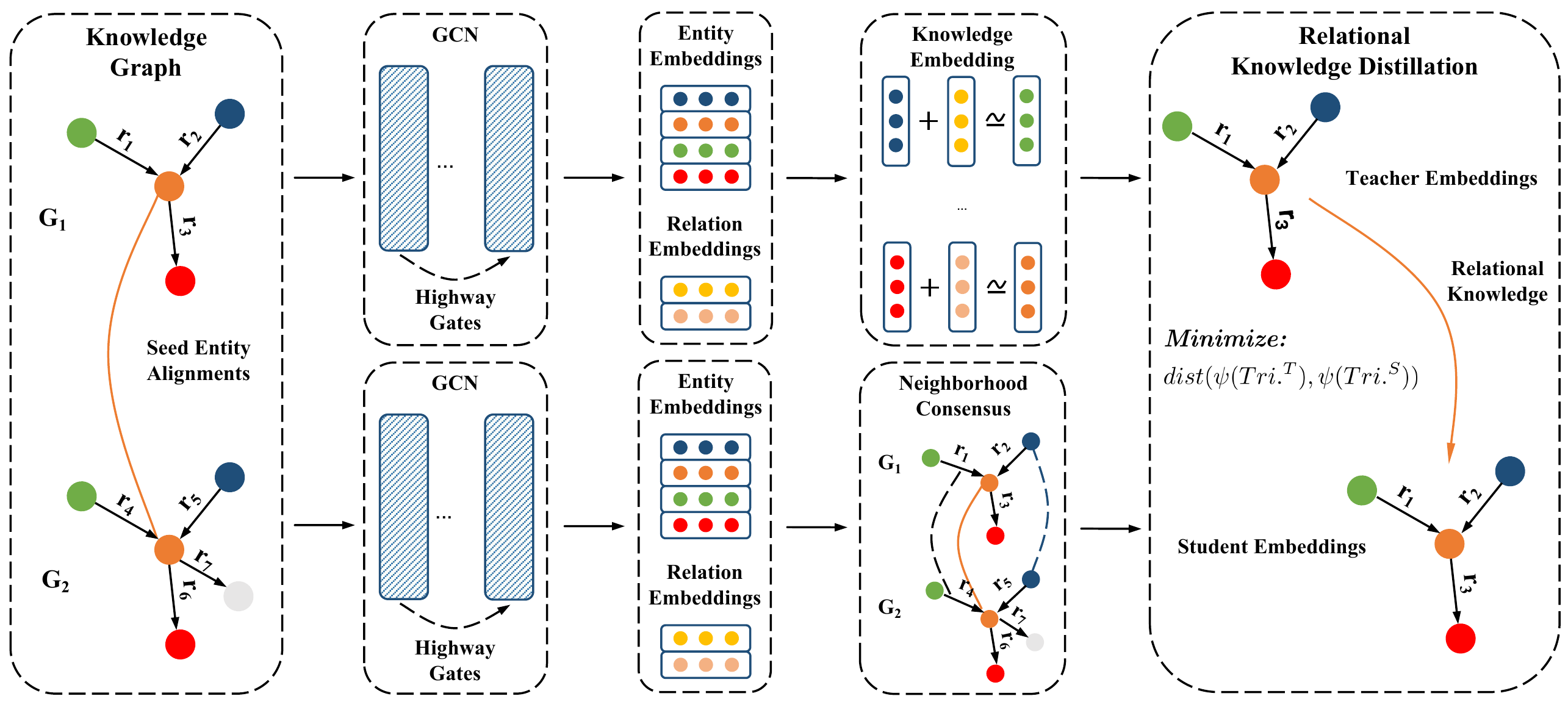} 
  \caption{Overall architecture: Relational Knowledge Distillation for Entity Alignment. The upper part is the knowledge embedding teacher model while the lower part is the neighborhood consensus student model. They have the similar GCN model structure with different training objectives. The relational knowledge of embeddings is transferred from teacher to student via relational knowledge distillation.}\label{fig-framework}
\end{figure*}

\section{Methodology}

In order to better utilize relational knowledge and neighborhood information, we propose a knowledge distillation based framework to consider knowledge embedding and neighborhood consensus simultaneously.

As shown in Figure~\ref{fig-framework}, our framework consists of three components:
\begin{compactitem}
\item A pre-trained two-layers GCN with highway gates as the \emph{teacher model} to provide relational knowledge, whose objective function is similar to TransE; 
\item A two-layers GCN with highway gates as the \emph{student model} to learn the local graph structure by neighborhood consensus with seed alignments;
\item A knowledge distillation mechanism to transfer relational knowledge from teacher model to student model, specifically an objective of minimizing distance-wise distillation loss.
\end{compactitem}

\subsection{Highway Gated GCN}

For both the teacher and student models, we utilize a GCN~\cite{DBLP:conf/iclr/KipfW17} based model to learn the representations of entities and relations.
Specifically, we use the highway gated GCN which could capture long-distance neighborhood information by stacking multiple GCN layers as the basic building block for our model.
The input of highway gated GCN model is an entity feature matrix $\mathbf{X} \in \mathbb{R}^{n \times d}$, where $n$ is the number of entities and $d$ is entity feature dimension.
For each GCN layer, the forward propagation is calculated as Equation~\eqref{eq-gcn}:
\begin{equation}\label{eq-gcn}
  \mathbf{H}^{l+1} =  \sigma (\tilde{D}^{-\frac{1}{2}} \tilde{A} \tilde{D}^{-\frac{1}{2}} \mathbf{H}^{l} \mathbf{W}^{l} )
\end{equation}  
where $\mathbf{H}^{l} \in \mathbb{R}^{n \times d^{l}}$ is the hidden state of the $l$-th GCN layer and $\mathbf{H}^{(0)} = \mathbf{X}$,
$\sigma(\cdot)$ is an activation function chosen as $\rm {ReLU}(\cdot) = max(0, \cdot)$, 
$\tilde{A} = A + I$ is an adjacency matrix derived from the connectivity matrix $A \in \mathbb{R}^{n \times n}$ of graph $G$ and an identity matrix $I \in \mathbb{R}^{n \times n}$ of self-connection,
$\tilde{D}$ denotes the diagonal node degree matrix of $\tilde{A}$,
$\mathbf{W}^{l} \in \mathbb{R}^{d^{l} \times d^{l+1}}$ and $d^{l}$ denote the weights and dimensions of features in layer $l$, respectively.

Following~\cite{DBLP:journals/corr/SrivastavaGS15}, we utilize layer-wise highway gates in forward propagation.
With the help of stacked GCN layers, rich neighborhood knowledge indicating graph structure information could be captured in learning entity embeddings.
The detailed calculating process is as Equation~\eqref{eq-hgcn1}: 
\begin{gather}\label{eq-hgcn1}
    T(\mathbf{H}^{l}) = \sigma (\mathbf{H}^{l} \mathbf{W}_{T}^{l} + \mathbf{b}_{T}^{l}) \\
   \mathbf{H}^{l+1} = T(\mathbf{H}^{l}) \cdot \mathbf{H}^{l+1} + (1-T(\mathbf{H}^{l})) \cdot \mathbf{H}^{l}
\end{gather}
\noindent where $\sigma$ is a sigmoid function;
$\mathbf{W}_T^{l}$ and $\mathbf{b}_T^{l}$ are weight matrix and bias vector of transform gate $T$, respectively;
$\cdot$ denotes element-wise multiplication;
$1-T$ represents the carry gate for vanilla input of each layer opposite to the transform gate $T$ for transformed input.

\subsection{Knowledge Embedding Model}

As shown in Figure~\ref{fig-framework}, the pre-trained teacher model aims at learning the knowledge embedding.
In this paper, we choose the objective function of TransE~\cite{DBLP:conf/nips/BordesUGWY13} as an example.
Note that it could also be replaced with other translation-based methods.
The relation triple is denoted as a translation equation $\mathbf{h} + \mathbf{r} \approx \mathbf{t}$, where $\mathbf{h, r, t}$ represents head entity, relation and tail entity, respectively.
For each triple $(e_h, r, e_t)$, we take $L_2$ normalized $f(e_h, r, e_t) = ||\mathbf{e_h} + \mathbf{r} - \mathbf{e_t}||_2$ as scoring function.
Following the previous studies, we apply negative sampling to generate negative unreal triples in the pre-training process.
The objective function of the knowledge embedding teacher model is shown as Equation~\eqref{eq-transe}:
\begin{equation}\label{eq-transe}
  \begin{split}
    \mathcal{L}_{KE}
    = \sum_{(e_h,r,e_t) \in T} \sum_{(e_h', r', e_t') \in T'} [||\mathbf{e}_h+\mathbf{r}-\mathbf{e}_t|| &\\
    - ||\mathbf{e}_h'+\mathbf{r}'-\mathbf{e}_t'|| +\gamma_1 ]_+ &
  \end{split}
\end{equation}
where $\rm [\cdot]_+ = max\{0, \cdot\}$, $e$ and $r$ are the embedding representations of entities and relations,
$T = T_1 \cup T_2$ denotes the aggregation of triples in two KGs, $G'$ represents the negative sampled triples set derived from $G$, $\gamma_1$ is a margin hyper-parameter with positive values.
For the sake of preserving semantic, we construct negative samples by randomly replacing the head or tail entity of an existing triple with $\mathcal{K}_1$ other entity with similar semantic.

\subsection{Neighborhood Consensus Model}

The neighborhood consensus student model has the similar structure to the knowledge embedding teacher model.
The only difference between them is the learning objective.
While the teacher model learns relational knowledge in triples, the student model learns local graph structure information in neighbors.
In order to calculate the neighborhood similarities between entities from different KGs, we utilize an \emph{energy function} of $L_2$ distance of neighborhood aggregated entity embeddings.
Specifically, given an entity pair $(e_1, e_2) \in E_1 \times E_2$, the similarity measure denotes as $d(e_1, e_2) = ||\mathbf{e_1} - \mathbf{e_2}||_2$.
The learning objective of the neighborhood consensus student model is to minimize the margin-based ranking loss in Equation~\eqref{eq-ea}:
\begin{equation}\label{eq-ea}
  \begin{split}
    \mathcal{L}_{NC} = \sum_{(e_1,e_2) \in \mathcal{S}_E} & \sum_{(e_1',e_2') \in \mathcal{S}_E'} [d(e_1, e_2) \\
    & - d(e_1', e_2')+\gamma_2]_+ 
  \end{split}
\end{equation}
where $\rm [\cdot]_+ = max \{ 0,\cdot \}$ denotes positive part of element,
$\gamma_2$ is a margin hyper-parameter,
$\mathcal{S}_E$ and $\mathcal{S}_E'$ represent sets of positive and negative entity pairs, respectively.
For the negative sampling, we choose the $\mathcal{K}_2$ nearest neighbors as the negative corresponding entities rather than random sampling.
Specifically, given an existing pair $(e_1, e_2)$, we replace $e_1$($e_2$) with the entity $e_1'$($e_2'$) that is closest to $e_2$($e_1$) on $L_2$ distance.

\subsection{Relational Knowledge Distillation}
To integrate relational knowledge and neighborhood information via knowledge distillation, we need to address two issues:
(i) how to learn the structural information along with contents via the distillation approach;
(ii) how to propose a learning objective that minimizes the difference between the teacher and student models.
Therefore, we need to dynamically adjust the contribution of knowledge embedding and neighborhood consensus during distillation.

To keep the relational knowledge in the process of distillation, we borrow the idea of \emph{energy function} proposed in~\cite{DBLP:conf/cvpr/ParkKLC19}.
The basic idea is that it first randomly samples $n$ instances from the training instances.
Next the energy function $\psi$ is applied to these instances to describe the relationship between them. 
Then the loss can be calculated as Equation~\eqref{eq-rkd}:
\begin{equation}\label{eq-rkd}
  \mathcal{L} = \sum_{(x_1, ..., x_n) \in \mathbf{X}^N} d(\psi(y_1^T, ..., y_n^T), \psi(y_1^S, ..., y_n^S))
\end{equation}
where $(x_1, ..., x_n) \in \mathbf{X}^N$ are $n$ randomly sampled training instances;
$d$ is the distance measure of potential relational knowledge between teacher and student models;
$y_i^T$ and $y_i^S$ are the output representations of input $x_i$ in teacher and student models, respectively.
With such a training loss, the relational knowledge could be kept in the process of distillation.

Following this formulation, we utilize the same L2 distance measure with the teacher model between head and tail entities in triples as energy function to keep the potential relational information as shown in Equation~\eqref{eq-energy-func}:
\begin{equation}\label{eq-energy-func}
  \psi(e_h, e_t) = \frac{1}{\mu}||\mathbf{e}_h - \mathbf{e}_t||_2
\end{equation}
where $\mu$ is a normalization factor to scale the distance of different vector spaces to the same scale.
We empirically define $\mu$ as Equation~\eqref{eq-mu}:
\begin{equation}\label{eq-mu}
  \mu = \frac{1}{N} \sum_{(e_h, r, e_t) \in T} ||\mathbf{e}_h - \mathbf{e}_t||_2
\end{equation}
where $T = T_1 \cup T_2$ represents all triples in two KGs and $N$ is the number of triples in $T$.

In order to improve the robustness of outliers, we propose to utilize Huber loss~\cite{DBLP:conf/ijcnn/KarasuyamaT10} rather than MSE loss as difference measure between the teacher and the student, which is shown as Equation~\eqref{eq-huber-loss}:
\begin{equation}\label{eq-huber-loss}
  \begin{split}
    d(x, y) = \left \{
    \begin{array}{lr}
      \frac{1}{2} (x-y)^2 ,  & |x-y| \leq 1,\\
      |x-y| - \frac{1}{2} ,  & otherwise.\\
    \end{array}
    \right.
  \end{split}
\end{equation}

Therefore, the objective of knowledge distillation $\mathcal{L}_{KD}$ is specified as Equation~\eqref{eq-rkd-obj}:
\begin{equation}\label{eq-rkd-obj}
  \mathcal{L}_{KD} = \sum_{(e_h, r, e_t) \in T} d(\psi(\mathbf{e}_h^T, \mathbf{e}_t^T), \psi(\mathbf{e}_h^S, \mathbf{e}_t^S))
\end{equation}
where $\mathbf{e^T}$ and $\mathbf{e^S}$ denote the embedding representations of the teacher and student model, respectively.

Next, we introduce how to dynamically adjust the contribution of alignment loss (for neighborhood consensus) and knowledge distillation loss (for knowledge embedding) in the learning objective of the student model.
It is controlled with the hyper-parameter \emph{Temperature} denoted as $\beta$.
As mentioned above, these two kinds of information may be adversarial due to different optimization directions.
To address this problem, $\beta$ should be dynamically adjusted during the training process.
Intuitively, in the early stage of training, it should focus on learning the relational knowledge where $\mathcal{L}_{KD}$ is more important; while in the late stage when two losses become very small, it should concentrate on the alignment loss to avoid overfitting to relational knowledge. 
Therefore, instead of using a static value of $\beta$, we set adaptive value of $\beta$ as shown in Equation~\eqref{eq-temp}:
\begin{equation}\label{eq-temp}
  \beta = |\mathcal{L}_{NC}| \times |\mathcal{L}_{KD}|
\end{equation}
where $|\cdot|$ denotes the loss value without gradient and $\beta \in (0, 1)$.

Therefore, the final learning objective of our model is shown as Equation~\eqref{eq-student}:
\begin{equation}\label{eq-student}
  \mathcal{L}_{Student} = (1-\beta) \mathcal{L}_{NC} + \beta \mathcal{L}_{KD}
\end{equation}
where $\mathcal{L}_{NC}$ denotes the neighborhood consensus alignment loss as Equation~\eqref{eq-ea} shows.

b\section{Experiment}


\subsection{Datasets}

Following many previous studies, we evaluate all methods on the popular DBP15K and DWY100K datasets.
\begin{compactitem}
\item {\textbf{DBP15K}}~\cite{DBLP:conf/semweb/SunHL17} is composed of three cross-lingual datasets derived from DBpedia representing three language pairs of KGs respectively, which are DBP$_\text{ZH-EN}$ (Chinese to English), DBP$_\text{JA-EN}$ (Japanese to English), and DBP$_\text{FR-EN}$ (French to English).
Each dataset consists of two KGs with hundreds of thousands of relation triples and 15K pre-aligned seed entity pairs along with seed relation pairs.

\item {\textbf{DWY100K}}~\cite{DBLP:conf/ijcai/SunHZQ18} contains two large-scale cross-domain datasets derived from DBpedia, Wikidata, and YAGO3, denoted as DWY$_\text{WD}$ (DBpedia to Wikidata) and DWY$_\text{YG}$ (DBpedia to YAGO3). 
Similar to DBP15K, DWY100K means 100K seed entity alignments in each dataset.
\end{compactitem}
The detailed statistics are shown in Table~\ref{tbl:datasets}.
In all experiments, we utilize 30\% of seed alignments in training, which is consistent with previous studies. 

\begin{table}
  \centering
  \scalebox{1.1}{
    \begin{tabular}{c|c|ccc}
      \toprule
      \multicolumn{2}{c|}{Datasets} & \#Ent. & \#Rel. & \#Rel. Triples \\
      \midrule
      \multirow{2}{*}{DBP$_\text{ZH-EN}$} & ZH & 66,469 & 2,830 & 153,929 \\
      & EN & 98,125 & 2,317 & 237,674 \\
      \midrule
      \multirow{2}{*}{DBP$_\text{JA-EN}$} & JA & 65,744 & 2,043 & 164,373 \\
      & EN & 95,680 & 2,096 & 233,319 \\
      \midrule
      \multirow{2}{*}{DBP$_\text{FR-EN}$} & FR & 66,858 & 1,379 & 192,191 \\
      & EN & 105,889 & 2,209 & 278,590 \\
      \midrule
      \multirow{2}{*}{DWY$_\text{WD}$} & DBpedia & 100,000 & 330 & 463,294 \\
      & Wikidata & 100,000 & 220 & 448,774 \\
      \midrule
      \multirow{2}{*}{DWY$_\text{YG}$} & DBpedia & 100,000 & 302 & 428,952 \\
      & YAGO3 & 100,000 & 31 & 502,563 \\
      \bottomrule
    \end{tabular}
  }
  \caption{Statistics of Datasets.}\label{tbl:datasets}
\end{table}

\subsection{Baselines}

To better verify the effectiveness of our proposed approach, we compare it with several state-of-the-art embedding-based models.
For knowledge embedding oriented methods, we choose \textbf{MTransE}~\cite{DBLP:conf/ijcai/ChenTYZ17}, \textbf{IPTransE}~\cite{DBLP:conf/ijcai/ZhuXLS17},
\textbf{SEA}~\cite{DBLP:conf/www/PeiYHZ19}, \textbf{RSN4EA}~\cite{DBLP:conf/icml/GuoSH19} as the representatives, which align entities based on relational knowledge.
For neighborhood consensus oriented methods, we choose  \textbf{GCN-Align}~\cite{DBLP:conf/emnlp/WangLLZ18}, \textbf{MuGNN}~\cite{DBLP:conf/acl/CaoLLLLC19}, \textbf{KECG}~\cite{DBLP:conf/emnlp/LiCHSLC19} and \textbf{AliNet}~\cite{DBLP:conf/aaai/SunW0CDZQ20}, which apply graph neural network to aggregate neighborhood information for alignment.
Among them, KECG explicitly models relational knowledge and neighborhood information with two learning objectives as our \name does.
These two methods are the most related ones to our proposed approach.

Since our work focuses on transferring relational knowledge rather than developing a better model for knowledge graphs entity alignment, we only utilize structure information for baselines for fair comparison according to the comprehensive survey~\cite{Zeng2021ACS}.
Although there are some other studies in this topic~\cite{DBLP:conf/ijcai/SunHZQ18,DBLP:conf/ijcai/Pei0Z19,DBLP:conf/wsdm/MaoWXLW20,DBLP:conf/aaai/YangLZWX20,DBLP:conf/ijcai/Tang0C00L20,DBLP:conf/emnlp/LiuCPLC20}, they mainly focus on utilizing other information, such as attribute and description or applying data enhancement, e.g. bootstraping strategy and machine translation. 
Such approaches are orthogonal to our work and they could also be enhanced on the basis of \name.
Therefore, we exclude the comparison with them here.

For ablation study, we design two variants of our \name, i.e., \name (w/o KD) that does not employ knowledge distillation, \name (w/o Temp.) that does not incorporate temperature factor to control the training process.

\subsection{Implementation Details}

In experiments, we choose the hyper-parameters by grid search as following:
Learning rate $\lambda$ is among \{0.0001, 0.0005, 0.001, 0.005, 0.01\}, $\gamma_1, \gamma_2$ are in \{1.0, 2.0, 3.0\}.
Specifically, the optimal values for these hyper-parameters are $\lambda = 0.005$ for knowledge embedding teacher model, $\lambda = 0.001$ for neighborhood consensus student model, $\gamma_1 = 3.0$, $\gamma_2 = 1.0$.
Following previous studies~\cite{DBLP:conf/ijcai/WuLF0Y019,DBLP:conf/emnlp/WuLFWZ19,DBLP:conf/acl/WuLFWZ20}, the dimensions of embedding vectors in both teacher model and student model are set to 300 and we use the pre-trained \textit{glove} of entity names as input features.
For DBP15K, the negative samples are updated each 50 epochs and the numbers of negative samples are set to $\mathcal{K}_1 = 10$ and $\mathcal{K}_2 = 200$, respectively.
For DWY100K, the negative samples are updated every 10 epochs and the numbers of negative samples are set to $\mathcal{K}_1 = 10$ and $\mathcal{K}_2 = 50$, respectively.
Following previous work, we use Hits@1 and Hits@10 as the main evaluation metrics.

\subsection{Results}

\begin{table*}[!ht]
  \centering
  \scalebox{1.1}{
    \begin{tabular}{l|ccc|ccc|ccc}
      \toprule
      \multirow{2}{*}{Models} & \multicolumn{3}{c|}{DBP$_\text{ZH-EN}$} & \multicolumn{3}{c|}{DBP$_\text{JA-EN}$} & \multicolumn{3}{c}{DBP$_\text{FR-EN}$}\\
      & Hits@1 & Hits@10 & MRR & Hits@1 & Hits@10 & MRR & Hits@1 & Hits@10 & MRR\\
      \midrule
      MTransE & 0.308 & 0.614 & 0.364 & 0.279 & 0.575 & 0.349 & 0.244 & 0.556 & 0.335\\
      IPTransE & 0.406 & 0.735 & 0.516 & 0.367 & 0.693 & 0.474 & 0.333 & 0.685 & 0.451\\
      SEA & 0.424 & ).796 & 0.548 & 0.385 & 0.783 & 0.518 & 0.400 & 0.797 & 0.533\\
      RSN4EA & 0.508 & 0.745 & 0.591 & 0.507 & 0.737 & 0.590 & 0.516 & 0.768 & 0.605\\
      \midrule
      GCN-Align & 0.413 & 0.744 & 0.549 & 0.399 & 0.745 & 0.546 & 0.373 & 0.745 & 0.532\\
      MuGNN & 0.494 & 0.844 & 0.611 & 0.501 & 0.857 & 0.621 & 0.495 & 0.870 & 0.621\\ 
      KECG & 0.478 & 0.835 & 0.598 & 0.490 & 0.844 & 0.610 & 0.486 & 0.851 & 0.610\\ 
      AliNet & 0.539 & 0.826 & 0.628 & 0.549 & 0.831 & 0.645 & 0.552 & 0.852 & 0.657\\ 
      \midrule
      (w/o RKD) & 0.438 & 0.802 & 0.564 & 0.462 & 0.811 & 0.574 & 0.446 & 0.822 & 0.583\\
      (w/o Temp.) & 0.573 & 0.857 & 0.677 & 0.576 & 0.873 & 0.681 & 0.564 & 0.862 & 0.673\\ 
      \name & \textbf{0.603} &\textbf{0.872} & \textbf{0.703} & \textbf{0.597} & \textbf{0.881} & \textbf{0.698} & \textbf{0.622} & \textbf{0.912} & \textbf{0.721}\\ 
      \bottomrule
    \end{tabular}
  }
  \caption{Performance Comparison on DBP15K datasets. The results are split into three parts by full lines. The upper part includes knowledge embedding methods; the middle part includes neighborhood consensus methods; while the bottom part includes three variants of our approach for the purpose of ablation study.}\label{tbl:results}
\end{table*}

\begin{table*}[!ht]
  \centering
  \scalebox{1.1}{
    \begin{tabular}{l|ccc|ccc}
      \toprule
      \multirow{2}{*}{Models} & \multicolumn{3}{c|}{DWY$_\text{WD}$} & \multicolumn{3}{c}{DWY$_\text{YG}$}\\
      & Hits@1 & Hits@10 & MRR & Hits@1 & Hits@10 & MRR \\
      \midrule
      MTransE & 0.281 & 0.520 & 0.363 & 0.252 & 0.493 & 0.334\\
      IPTransE & 0.349 & 0.638 & 0.447 & 0.297 & 0.558 & 0.386\\
      SEA & 0.518 & 0.802 & 0.616 & 0.516 & 0.736 & 0.592\\
      RSN4EA & 0.607 & 0.793 & 0.673 & 0.689 & 0.878 & 0.756\\
      \midrule
      GCN-Align & 0.506 & 0.772 & 0.600 & 0.597 & 0.838 & 0.682\\ 
      MuGNN & 0.616 & 0.897 & 0.714 & 0.741 & 0.937 & 0.810\\ 
      KECG & 0.632 & 0.900 & 0.728 & 0.728 & 0.915 & 0.798\\ 
      AliNet & 0.690 & 0.908 & 0.766 & 0.786 & 0.943 & 0.841\\ 
      \midrule
      (w/o RKD) & 0.577 & 0.848 & 0.659 & 0.671 & 0.889 & 0.751\\ 
      (w/o Temp.) & 0.703 & 0.921 & 0.773 & 0.818 & 0.961 & 0.870\\ 
      \name & \textbf{0.756} & \textbf{0.973} & \textbf{0.821} & \textbf{0.823} & \textbf{0.971} & \textbf{0.879}\\ 
      \bottomrule
    \end{tabular}
  }
  \caption{Performance Comparison on DWY100K datasets. The results are split into three parts by full lines. The upper part includes knowledge embedding methods; the middle part includes neighborhood consensus methods; while the bottom part includes three variants of our approach for the purpose of ablation study.}\label{tbl:results2}
\end{table*}

Table~\ref{tbl:results} shows the results of experiments on DBP15K and DWY100K.
It can be seen that \name achieves promising performance on both cross-lingual and cross-domain datasets, indicating the effectiveness of our proposed framework.
Moreover, \name achieves significant improvement over the compared baseline methods on the DBP15K dataset.
The reason is that those baseline methods fail to incorporate complex relational knowledge due to the sparsity of DBP15K while our \name is capable of exploiting fine-grained relational knowledge.
Although KECG and HyperKA methods also explicitly learn knowledge embedding and neighborhood consensus, they fail to propose an effective way to integrate two different objectives.
Meanwhile, with the help of knowledge distillation, \name can effectively and flexibly incorporate relational knowledge into neighborhood consensus model and thus achieves much better performance.

In the large-scale dataset DWY100K, \name also significantly outperforms all other methods.
Since DWY100K is much larger than DBP15K with fewer relations and the graph structure is more similar, the neighborhood consensus plays a more important role, and the performance gain by utilizing knowledge distillation is less than DBP15K.
Even though, \name still reports reasonable results due to the properly designed techniques.

\subsection{Effectiveness of Knowledge Distillation}

\begin{table}
  \centering
  \scalebox{1}{
    \begin{tabular}{l|ccc|ccc|ccc}
      \toprule
      \multirow{2}{*}{Models} & \multicolumn{3}{c|}{DBP$_\text{ZH-EN}$} & \multicolumn{3}{c|}{DBP$_\text{JA-EN}$} & \multicolumn{3}{c}{DBP$_\text{FR-EN}$} \\
      & Hits@1 & Hits@10 & MRR & Hits@1 & Hits@10 & MRR & Hits@1 & Hits@10 & MRR\\
      \midrule
      KECG (w/o KE) & 0.430 & 0.791 & 0.551 & 0.446 & 0.807 & 0.567 & 0.432 & 0.815 & 0.559 \\
      KECG (w/ Init.) & 0.481 & 0.823 & 0.589 & 0.473 & 0.823 & 0.583 & 0.461 & 0.826 & 0.574 \\
      KECG & 0.478 & 0.835 & 0.598 & 0.490 & 0.844 & 0.610 & 0.486 & 0.851 & 0.610 \\
      KECG (w/ KD) & \textbf{0.513} & \textbf{0.853} & \textbf{0.627} & \textbf{0.516} & \textbf{0.861} & \textbf{0.633} & \textbf{0.535} & \textbf{0.877} & \textbf{0.651} \\
      \midrule
      HyperKA (w/o KE) & 0.518 & 0.814 & 0.623 & 0.535 & 0.834 & 0.640 & 0.529 & 0.859 & 0.645 \\
      HyperKA (w/ Init.) & 0.569 & 0.847 & 0.659 & 0.551 & 0.853 & 0.659 & 0.572 & 0.878 & 0.681 \\
      HyperKA & 0.572 & 0.865 & 0.678 & 0.564 & 0.865 & 0.673 & 0.597 & 0.891 & 0.704 \\
      HyperKA (w/ KD) & \textbf{0.581} & \textbf{0.868} & \textbf{0.693} & \textbf{0.584} & \textbf{0.879} & \textbf{0.691} & \textbf{0.601} & \textbf{0.894} & \textbf{0.711} \\
      \bottomrule
    \end{tabular}
  }
  \caption{Results of KECG and HyperKA with different knowledge embedding methods on DBP15K.}\label{tbl:kd}
\end{table}

To further analyze the importance of relational knowledge and the effectiveness of relational knowledge distillation, we integrate our knowledge distillation techniques with KECG and HyperKA models, producing four variants with different knowledge embedding methods for each model. Specifically, (w/o KE) denotes variants without knowledge embedding, (w/ Init.) denotes variants with entity embedding initialization of pre-trained knowledge embedding model and (w/ RKD) denotes variants with our relational knowledge distillation. As shown in Table~\ref{tbl:kd}, our distillation method (w/ RKD) yields the best performances for both KECG and HyperKA.

\begin{figure}
  \centering
  \includegraphics[width=\linewidth]{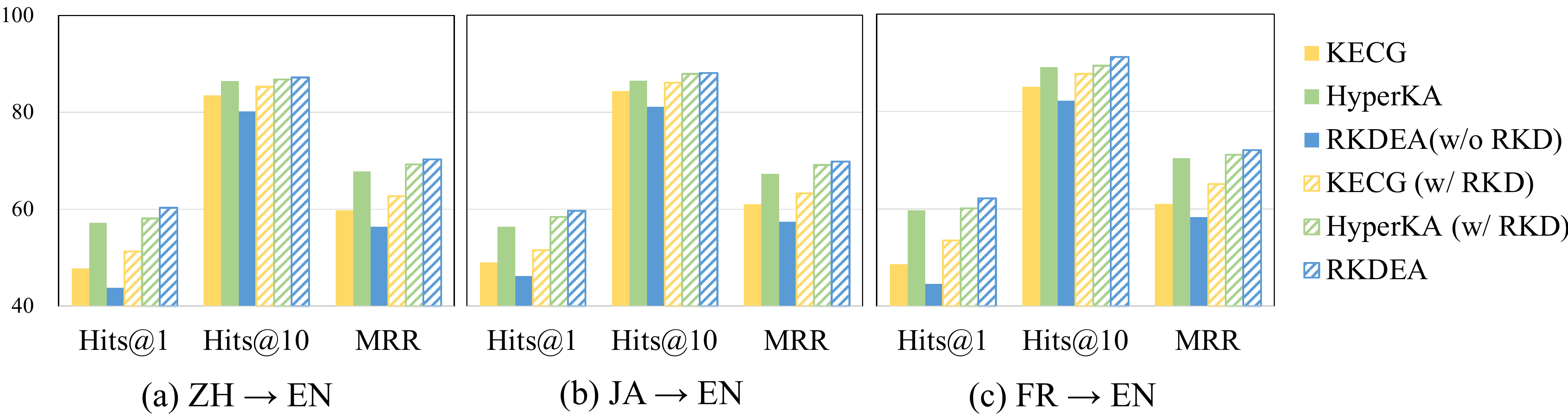}
  \caption{Effectiveness evaluation of incorporating Relational Knowledge Distillation (RKD) on DBP15K. The solid filled columns indicate models without RKD while the slash filled columns indicate models with RKD.}\label{fig:kd}
\end{figure}

In order to illustrate the performance gain brought by our proposed knowledge distillation approach, Figure~\ref{fig:kd} shows the performance comparison among the models with and without relational knowledge distillation on DBP15K.
The results clearly show that by introducing relational knowledge distillation, all three models, KECG, HyperKA and RKDEA achieve significant performance gain.

These results demonstrate that our proposed methods could be adopted to improve other existing KG alignment models and therefore further prove the potential and effectiveness of our proposed relational knowledge distillation method.

\subsection{Impact of Adaptive Temperature Factor}
\begin{figure}
  \centering
  \includegraphics[width=\linewidth]{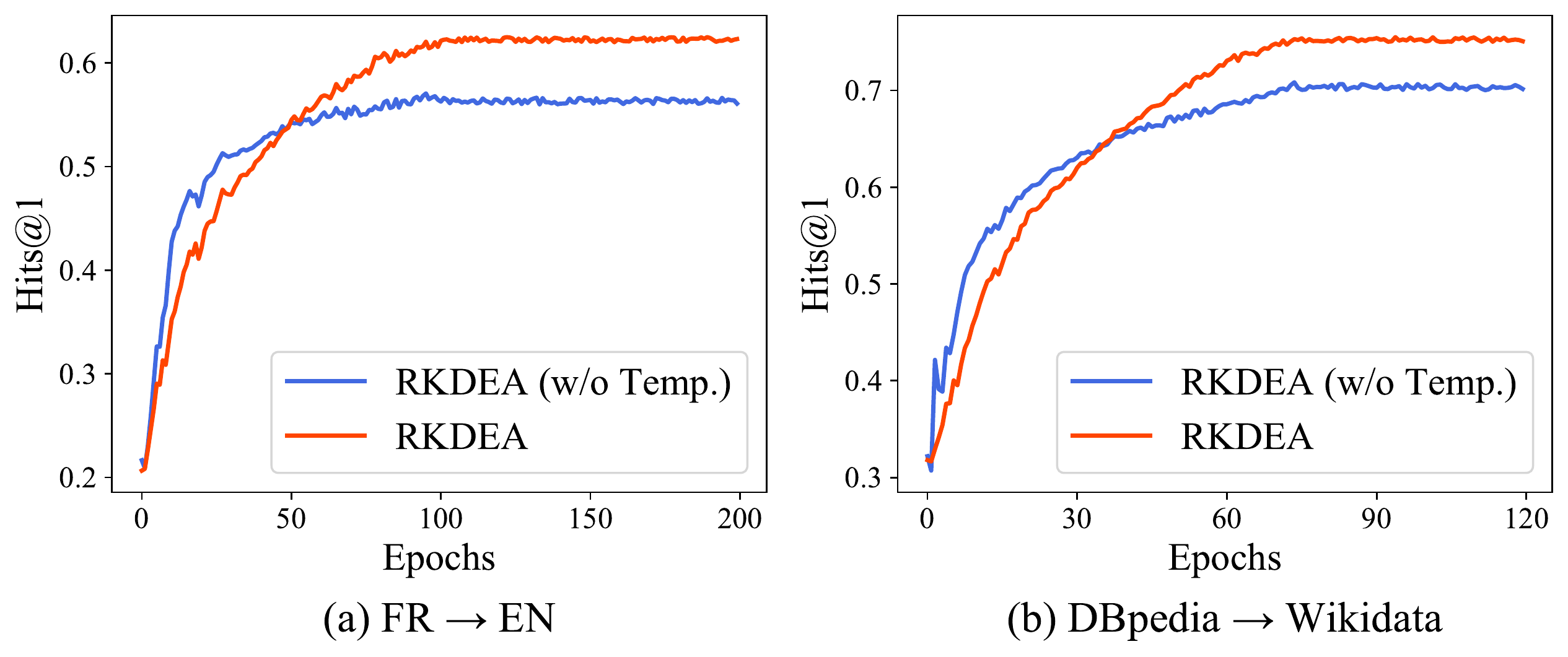}
  \caption{Impact of temperature factor in training process on DBP$_\text{FR-EN}$ and DWY$_\text{WD}$.}\label{fig:temp}
\end{figure}

The adaptive temperature mechanism is one of the core contributions in \name.
To explore the effect of incorporating the temperature factor, we conduct ablation study by comparing with \name (w/o Temp.) on both cross-lingual and cross-domain datasets.
Figure~\ref{fig:temp} shows the comparison of Hits@1 change curve with the iteration increasing between these two models during the training process.
The results illustrate that \name (w/o Temp.) converges faster while \name achieves higher Hits@1 at the end of training process.

Actually, the temperature is a weight decay mechanism in the training process.
In the early stage of training, when GNN is not well trained, the distilled relational knowledge is instructive for entity alignment and makes the model converge quickly into a relatively good state.
However, as the training progress moves on, relational knowledge and neighborhood information may lead to different objectives.
Therefore, if the contribution of distilled relational knowledge stays the same, the model will fall into \textit{trade-off} between two directions:
\textit{overfit} the relational knowledge but \textit{underfit} the neighborhood consensus or \textit{overfit} the neighborhood consensus but \textit{underfit} the relational knowledge.
Consequently, involving excessive information could be potentially harmful while the adaptive temperature mechanism can avoid it by controlling the contribution of distilled relational knowledge.

\section{Conclusion}
In this paper, we study the problem of entity alignment over heterogeneous KGs.
We propose a GCN based framework with knowledge distillation techniques to take advantage of the complex relational knowledge by jointly learning entity embedding and neighborhood consensus.
With the help of relational knowledge distillation, our model can effectively and flexibly model relational knowledge and neighborhood information.
Furthermore, by automatically adjusting the temperature parameter, our proposed model can dynamically control the contribution of different objectives and avoid overfitting.
Experimental results on several popular benchmarking datasets show that the proposed solutions outperform the state-of-the-art methods by an obvious margin.

%
%
%

\begin{thebibliography}{10}
\providecommand{\url}[1]{\texttt{#1}}
\providecommand{\urlprefix}{URL }
\providecommand{\doi}[1]{https://doi.org/#1}

\bibitem{DBLP:conf/nips/BaC14}
Ba, J., Caruana, R.: In: NeurIPS. pp. 2654--2662 (2014)

\bibitem{DBLP:journals/corr/abs-1805-02641}
Bagherinezhad, H., Horton, M., Rastegari, M., Farhadi, A.: Label refinery:
  Improving imagenet classification through label progression. CoRR
  \textbf{abs/1805.02641} (2018)

\bibitem{DBLP:conf/nips/BordesUGWY13}
Bordes, A., Usunier, N., Garc{\'{\i}}a{-}Dur{\'{a}}n, A., Weston, J.,
  Yakhnenko, O.: Translating embeddings for modeling multi-relational data. In:
  NeurIPS. pp. 2787--2795 (2013)

\bibitem{DBLP:conf/kdd/BucilaCN06}
Bucila, C., Caruana, R., Niculescu{-}Mizil, A.: Model compression. In: SIGKDD.
  pp. 535--541 (2006)

\bibitem{DBLP:conf/acl/CaoLLLLC19}
Cao, Y., Liu, Z., Li, C., Liu, Z., Li, J., Chua, T.: Multi-channel graph neural
  network for entity alignment. In: ACL. pp. 1452--1461 (2019)

\bibitem{DBLP:conf/pakdd/ChenZT0L20}
Chen, B., Zhang, J., Tang, X., Chen, H., Li, C.: Jarka: Modeling attribute
  interactions for cross-lingual knowledge alignment. In: PAKDD. vol. 12084,
  pp. 845--856 (2020)

\bibitem{DBLP:conf/ijcai/ChenTYZ17}
Chen, M., Tian, Y., Yang, M., Zaniolo, C.: Multilingual knowledge graph
  embeddings for cross-lingual knowledge alignment. In: IJCAI. pp. 1511--1517
  (2017)

\bibitem{DBLP:conf/aaai/ChenWZ18}
Chen, Y., Wang, N., Zhang, Z.: Darkrank: Accelerating deep metric learning via
  cross sample similarities transfer. In: McIlraith, S.A., Weinberger, K.Q.
  (eds.) AAAI. pp. 2852--2859 (2018)

\bibitem{DBLP:conf/icml/FurlanelloLTIA18}
Furlanello, T., Lipton, Z.C., Tschannen, M., Itti, L., Anandkumar, A.:
  Born-again neural networks. In: ICML. vol.~80, pp. 1602--1611 (2018)

\bibitem{DBLP:conf/icml/GuoSH19}
Guo, L., Sun, Z., Hu, W.: Learning to exploit long-term relational dependencies
  in knowledge graphs. In: ICML. vol.~97, pp. 2505--2514 (2019)

\bibitem{DBLP:conf/ccks/HaoZHL016}
Hao, Y., Zhang, Y., He, S., Liu, K., Zhao, J.: A joint embedding method for
  entity alignment of knowledge bases. In: CCKS. vol.~650, pp. 3--14 (2016)

\bibitem{DBLP:journals/corr/HintonVD15}
Hinton, G.E., Vinyals, O., Dean, J.: Distilling the knowledge in a neural
  network  (2015)

\bibitem{DBLP:journals/corr/HuangW17a}
Huang, Z., Wang, N.: Like what you like: Knowledge distill via neuron
  selectivity transfer. CoRR  \textbf{abs/1707.01219} (2017)

\bibitem{DBLP:conf/acl/JiHXL015}
Ji, G., He, S., Xu, L., Liu, K., Zhao, J.: Knowledge graph embedding via
  dynamic mapping matrix. In: ACL. pp. 687--696 (2015)

\bibitem{DBLP:conf/ijcnn/KarasuyamaT10}
Karasuyama, M., Takeuchi, I.: Nonlinear regularization path for the modified
  huber loss support vector machines. In: IJCNN. pp.~1--8 (2010)

\bibitem{DBLP:conf/iclr/KipfW17}
Kipf, T.N., Welling, M.: Semi-supervised classification with graph
  convolutional networks. In: ICLR (2017)

\bibitem{DBLP:conf/aaai/LeeKLY17}
Lee, J., Kim, H., Lee, J., Yoon, S.: Transfer learning for deep learning on
  graph-structured data. In: AAAI. pp. 2154--2160 (2017)

\bibitem{DBLP:conf/emnlp/LiCHSLC19}
Li, C., Cao, Y., Hou, L., Shi, J., Li, J., Chua, T.: Semi-supervised entity
  alignment via joint knowledge embedding model and cross-graph model. In:
  EMNLP. pp. 2723--2732 (2019)

\bibitem{DBLP:journals/jdiq/LiLSWHT21}
Li, Y., Li, J., Suhara, Y., Wang, J., Hirota, W., Tan, W.: Deep entity
  matching: Challenges and opportunities. {ACM} J. Data Inf. Qual.
  \textbf{13}(1),  1:1--1:17 (2021)

\bibitem{DBLP:conf/aaai/LinLSLZ15}
Lin, Y., Liu, Z., Sun, M., Liu, Y., Zhu, X.: Learning entity and relation
  embeddings for knowledge graph completion. In: AAAI. pp. 2181--2187 (2015)

\bibitem{DBLP:journals/tkde/LiuADZWH22}
Liu, Y., Ao, X., Dong, L., Zhang, C., Wang, J., He, Q.: Spatiotemporal activity
  modeling via hierarchical cross-modal embedding. {IEEE} Trans. Knowl. Data
  Eng.  \textbf{34}(1),  462--474 (2022)

\bibitem{DBLP:conf/emnlp/LiuCPLC20}
Liu, Z., Cao, Y., Pan, L., Li, J., Chua, T.: Exploring and evaluating
  attributes, values, and structures for entity alignment. In: EMNLP. pp.
  6355--6364 (2020)

\bibitem{DBLP:conf/cikm/LuLW019}
Lu, J., Lin, C., Wang, J., Li, C.: Synergy of database techniques and machine
  learning models for string similarity search and join. In: CIKM. pp.
  2975--2976 (2019)

\bibitem{DBLP:conf/cidr/MahdisoltaniBS15}
Mahdisoltani, F., Biega, J., Suchanek, F.M.: {YAGO3:} {A} knowledge base from
  multilingual wikipedias. In: CIDR (2015)

\bibitem{DBLP:conf/wsdm/MaoWXLW20}
Mao, X., Wang, W., Xu, H., Lan, M., Wu, Y.: {MRAEA:} an efficient and robust
  entity alignment approach for cross-lingual knowledge graph. In: WSDM. pp.
  420--428 (2020)

\bibitem{DBLP:conf/ijcai/NieHSWCWZ20}
Nie, H., Han, X., Sun, L., Wong, C.M., Chen, Q., Wu, S., Zhang, W.: Global
  structure and local semantics-preserved embeddings for entity alignment. In:
  IJCAI. pp. 3658--3664 (2020)

\bibitem{DBLP:conf/cvpr/ParkKLC19}
Park, W., Kim, D., Lu, Y., Cho, M.: Relational knowledge distillation. In:
  CVPR. pp. 3967--3976 (2019)

\bibitem{DBLP:conf/www/PeiYHZ19}
Pei, S., Yu, L., Hoehndorf, R., Zhang, X.: Semi-supervised entity alignment via
  knowledge graph embedding with awareness of degree difference. In: WWW. pp.
  3130--3136 (2019)

\bibitem{DBLP:conf/ijcai/Pei0Z19}
Pei, S., Yu, L., Zhang, X.: Improving cross-lingual entity alignment via
  optimal transport. In: IJCAI. pp. 3231--3237 (2019)

\bibitem{DBLP:conf/nips/RoccoCATPS18}
Rocco, I., Cimpoi, M., Arandjelovic, R., Torii, A., Pajdla, T., Sivic, J.:
  Neighbourhood consensus networks. In: NeurIPS. pp. 1658--1669 (2018)

\bibitem{DBLP:journals/corr/RomeroBKCGB14}
Romero, A., Ballas, N., Kahou, S.E., Chassang, A., Gatta, C., Bengio, Y.:
  Fitnets: Hints for thin deep nets. In: ICLR (2015)

\bibitem{DBLP:conf/esws/SchlichtkrullKB18}
Schlichtkrull, M.S., Kipf, T.N., Bloem, P., van~den Berg, R., Titov, I.,
  Welling, M.: Modeling relational data with graph convolutional networks. In:
  ESWC. pp. 593--607 (2018)

\bibitem{DBLP:journals/corr/SrivastavaGS15}
Srivastava, R.K., Greff, K., Schmidhuber, J.: Highway networks  (2015)

\bibitem{DBLP:conf/semweb/SunHL17}
Sun, Z., Hu, W., Li, C.: Cross-lingual entity alignment via joint
  attribute-preserving embedding. In: ISWC. vol. 10587, pp. 628--644 (2017)

\bibitem{DBLP:conf/ijcai/SunHZQ18}
Sun, Z., Hu, W., Zhang, Q., Qu, Y.: Bootstrapping entity alignment with
  knowledge graph embedding. In: IJCAI. pp. 4396--4402 (2018)

\bibitem{DBLP:conf/aaai/SunW0CDZQ20}
Sun, Z., Wang, C., Hu, W., Chen, M., Dai, J., Zhang, W., Qu, Y.: Knowledge
  graph alignment network with gated multi-hop neighborhood aggregation. In:
  AAAI. pp. 222--229 (2020)

\bibitem{DBLP:conf/ijcai/Tang0C00L20}
Tang, X., Zhang, J., Chen, B., Yang, Y., Chen, H., Li, C.: {BERT-INT:} {A}
  bert-based interaction model for knowledge graph alignment. In: IJCAI. pp.
  3174--3180 (2020)

\bibitem{DBLP:conf/nips/TarvainenV17}
Tarvainen, A., Valpola, H.: Mean teachers are better role models:
  Weight-averaged consistency targets improve semi-supervised deep learning
  results. In: NeurIPS. pp. 1195--1204 (2017)

\bibitem{DBLP:conf/ijcai/TianZWX19}
Tian, B., Zhang, Y., Wang, J., Xing, C.: Hierarchical inter-attention network
  for document classification with multi-task learning. In: IJCAI. pp.
  3569--3575 (2019)

\bibitem{DBLP:journals/isci/WangLLZ20}
Wang, J., Lin, C., Li, M., Zaniolo, C.: Boosting approximate dictionary-based
  entity extraction with synonyms. Inf. Sci.  \textbf{530},  1--21 (2020)

\bibitem{DBLP:conf/icde/WangLZ19}
Wang, J., Lin, C., Zaniolo, C.: Mf-join: Efficient fuzzy string similarity join
  with multi-level filtering. In: ICDE. pp. 386--397 (2019)

\bibitem{DBLP:conf/aaai/WangZFC14}
Wang, Z., Zhang, J., Feng, J., Chen, Z.: Knowledge graph embedding by
  translating on hyperplanes. In: AAAI. pp. 1112--1119 (2014)

\bibitem{DBLP:conf/ijcai/WangLT13}
Wang, Z., Li, J., Tang, J.: Boosting cross-lingual knowledge linking via
  concept annotation. In: IJCAI. pp. 2733--2739 (2013)

\bibitem{DBLP:conf/emnlp/WangLLZ18}
Wang, Z., Lv, Q., Lan, X., Zhang, Y.: Cross-lingual knowledge graph alignment
  via graph convolutional networks. In: EMNLP. pp. 349--357 (2018)

\bibitem{DBLP:conf/icde/WuZWLFX19}
Wu, J., Zhang, Y., Wang, J., Lin, C., Fu, Y., Xing, C.: Scalable metric
  similarity join using mapreduce. In: ICDE. pp. 1662--1665 (2019)

\bibitem{DBLP:conf/ijcai/WuLF0Y019}
Wu, Y., Liu, X., Feng, Y., Wang, Z., Yan, R., Zhao, D.: Relation-aware entity
  alignment for heterogeneous knowledge graphs. In: IJCAI. pp. 5278--5284
  (2019)

\bibitem{DBLP:conf/emnlp/WuLFWZ19}
Wu, Y., Liu, X., Feng, Y., Wang, Z., Zhao, D.: Jointly learning entity and
  relation representations for entity alignment. In: EMNLP. pp. 240--249 (2019)

\bibitem{DBLP:conf/acl/WuLFWZ20}
Wu, Y., Liu, X., Feng, Y., Wang, Z., Zhao, D.: Neighborhood matching network
  for entity alignment. In: ACL. pp. 6477--6487 (2020)

\bibitem{DBLP:conf/acl/XuWYFSWY19}
Xu, K., Wang, L., Yu, M., Feng, Y., Song, Y., Wang, Z., Yu, D.: Cross-lingual
  knowledge graph alignment via graph matching neural network. In: ACL. pp.
  3156--3161 (2019)

\bibitem{DBLP:conf/icde/YangZZWHX19}
Yang, J., Zhang, Y., Zhou, X., Wang, J., Hu, H., Xing, C.: A hierarchical
  framework for top-k location-aware error-tolerant keyword search. In: ICDE.
  pp. 986--997 (2019)

\bibitem{DBLP:conf/aaai/YangLZWX20}
Yang, K., Liu, S., Zhao, J., Wang, Y., Xie, B.: {COTSAE:} co-training of
  structure and attribute embeddings for entity alignment. In: AAAI. pp.
  3025--3032 (2020)

\bibitem{DBLP:conf/ijcai/Ye0FZW19}
Ye, R., Li, X., Fang, Y., Zang, H., Wang, M.: A vectorized relational graph
  convolutional network for multi-relational network alignment. In: IJCAI. pp.
  4135--4141 (2019)

\bibitem{DBLP:conf/cvpr/YimJBK17}
Yim, J., Joo, D., Bae, J., Kim, J.: A gift from knowledge distillation: Fast
  optimization, network minimization and transfer learning. In: CVPR. pp.
  7130--7138 (2017)

\bibitem{DBLP:conf/iclr/ZagoruykoK17}
Zagoruyko, S., Komodakis, N.: Paying more attention to attention: Improving the
  performance of convolutional neural networks via attention transfer. In: ICLR
  (2017)

\bibitem{Zeng2021ACS}
Zeng, K., Li, C., Hou, L., Li, J.Z., Feng, L.: A comprehensive survey of entity
  alignment for knowledge graphs. AI Open  \textbf{2},  1--13 (2021)

\bibitem{DBLP:conf/ijcai/ZhangSHCGQ19}
Zhang, Q., Sun, Z., Hu, W., Chen, M., Guo, L., Qu, Y.: Multi-view knowledge
  graph embedding for entity alignment. In: IJCAI. pp. 5429--5435 (2019)

\bibitem{DBLP:journals/www/ZhangCYWHXZ21}
Zhang, Y., Chen, Y., Yang, J., Wang, J., Hu, H., Xing, C., Zhou, X.: Clustering
  enhanced error-tolerant top-k spatio-textual search. World Wide Web
  \textbf{24}(4),  1185--1214 (2021)

\bibitem{DBLP:journals/tkde/ZhangWWX20}
Zhang, Y., Wu, J., Wang, J., Xing, C.: A transformation-based framework for
  {KNN} set similarity search. {IEEE} Trans. Knowl. Data Eng.  \textbf{32}(3),
  409--423 (2020)

\bibitem{DBLP:conf/dasfaa/ZhaoZWYZWX18}
Zhao, K., Zhang, Y., Wang, Z., Yin, H., Zhou, X., Wang, J., Xing, C.: Modeling
  patient visit using electronic medical records for cost profile estimation.
  In: Pei, J., Manolopoulos, Y., Sadiq, S.W., Li, J. (eds.) DASFAA. pp. 20--36
  (2018)

\bibitem{DBLP:conf/ijcai/ZhuXLS17}
Zhu, H., Xie, R., Liu, Z., Sun, M.: Iterative entity alignment via joint
  knowledge embeddings. In: IJCAI. pp. 4258--4264 (2017)

\bibitem{DBLP:conf/ijcai/ZhuZ0TG19}
Zhu, Q., Zhou, X., Wu, J., Tan, J., Guo, L.: Neighborhood-aware attentional
  representation for multilingual knowledge graphs. In: IJCAI. pp. 1943--1949
  (2019)

\end{thebibliography}

\end{document}